# Disruptive Event Classification using PMU Data in Distribution Networks

I. Niazazari and H. Livani, *Member, IEEE*

*Abstract*— Proliferation of advanced metering devices with high sampling rates in distribution grids, e.g., micro-phasor measurement units (µPMU), provides unprecedented potentials for wide-area monitoring and diagnostic applications, e.g., situational awareness, health monitoring of distribution assets. Unexpected disruptive events interrupting the normal operation of assets in distribution grids can eventually lead to permanent failure with expensive replacement cost over time. Therefore, disruptive event classification provides useful information for preventive maintenance of the assets in distribution networks. Preventive maintenance provides wide range of benefits in terms of time, avoiding unexpected outages, maintenance crew utilization, and equipment replacement cost. In this paper, a PMU-data-driven framework is proposed for classification of disruptive events in distribution networks. The two disruptive events, i.e., malfunctioned capacitor bank switching and malfunctioned regulator on-load tap changer (OLTC) switching are considered and distinguished from the normal abrupt load change in distribution grids. The performance of the proposed framework is verified using the simulation of the events in the IEEE 13-bus distribution network. The event classification is formulated using two different algorithms as; i) principle component analysis (PCA) together with multi-class support vector machine (SVM), and ii) autoencoder along with softmax classifier. The results demonstrate the effectiveness of the proposed algorithms and satisfactory classification accuracies.

*Index Terms*—Classification, PMU, Event Detection, SVM

## I. INTRODUCTION

With the advent of advanced metering devices, e.g., phasor measurement units (PMUs), and micro-PMUs (µPMUs), power transmission and distribution networks have become more intelligent, reliable and efficient [1], [2]. Most of the early measurement devices have had the limited measuring capacity, while the multi-functional capabilities of PMUs have made them important monitoring assets to power networks. PMUs have been used for several monitoring and control applications in transmission grids, e.g., state estimation [3], dynamic stability assessment [4], event diagnostics and classification [5]. Recently, the use of PMUs and µPMUs for several monitoring and control applications in distribution networks have been introduced. In [6], a fault location algorithm in distribution networks have been proposed using PMU data. In [7], PMUs are utilized in distribution networks with distributed generation for three different applications, namely, state estimation, protection and instability prediction. In [8], PMUs are deployed in distribution grids for measurement of synchronized harmonic phasors. Additionally, PMUs can help power grids to restore quicker in case of cutting off the energy supply by providing voltage, current and frequency measurements for reclosing the circuit breakers, e.g., 2008 Florida blackout [9]. In [10], PMU deployment for state estimation in active distribution networks is discussed. Reference [11] presents the use of PMU data for abnormal situation detection in distribution networks.

Event detection is an ongoing field of study, and can be a challenging task due to different types of correlated events in distribution grids, e.g., switching vs. load changing. Therefore, distinguishing the disruptive events from one another, and differentiating them from a normal condition of the network, requires advanced data-driven frameworks. Several different techniques are proposed in the literature for classification of events in power networks. In [12], a probabilistic neural network along with S-transform is utilized to classify power quality disturbances. Partial discharge pattern recognition is conducted by applying fuzzy decision tree method [13], and sparse representation and artificial neural network [14].

Support vector machine (SVM) is a widely used method for event classification, e.g., fault location [15], power system security assessment [16], and transient stability analysis [17].

Accurate distribution event detection and classification results in accurate preventive maintenance scheduling of the critical assets based on the warning signs of the pending electrical failures. Preventive maintenance is a beneficial task in terms of time, equipment replacement cost, maintenance crew utilization, avoiding unexpected outages, and consequently, extending the life of the critical assets. Real-time data analytics can help to detect multiple failures, along with offering online monitoring of feeder operations. Therefore, it can provide utilities with useful information about faulty equipment in particular parts of the network. In [18], the authors have used highly sensitive waveform recorders for gathering data and improving feeders' visibility and operational efficiency. The collected data from waveform recorders is used for incipient equipment failures detection on distribution feeders [19]. In [20], a data-driven methodology is presented for classification of five disruptive events, i.e., cable failure, hot-line clamp failure, vegetation intrusion resulting in frequent fault, fault induced conductor slap, and capacitor controller malfunction.

In this paper, we propose a framework for classification of two disruptive events from the normal abrupt load change in distribution networks using PMU data. These classes are malfunctioned capacitor bank switching, malfunctioned regulator on-load tap changer (OLTC) switching, and abrupt load changing. The disruptive events may not cause immediate failure, however, they will cause permanent equipment failure and expensive replacement cost over time. The classification of these events prioritizes the preventive maintenance scheduling and leads to higher life expectancy of distribution assets. In this paper, the classification algorithms are developed using (1) principal component analysis (PCA) along with SVM, and (2) autoencoder along with softmax layer.

The rest of this paper is organized as follows, in section II, the problem statement and the proposed methodology is presented. Section III presents the simulation results, and the conclusion and future works are presented in section IV.

## II. PROBLEM STATEMENT AND METHODOLOGY

Data analytics plays a major role in power system post-event analysis such as event diagnosis and preventive maintenance scheduling. These applications are helpful in terms of saving maintenance time and cost, and leads to preventing unexpected outages due to permanent failure of critical assets. In this paper, two different disruptive events, i.e., malfunctioned capacitor bank switching and malfunctioned regulator OLTC switching, along with a normal abrupt load changing, are categorized as three different classes. Malfunctioned capacitor bank switching events are caused by failure in mechanical switches and can happen in less than 2 cycles, i.e., 32 *msec*. The malfunctioned regulator OLTC switching can be caused due to ageing and degradation of the selector switches. In a malfunctioned regulator OLTC switching, the tap is dislocated, and after a while relocated to its original position. In this paper, we propose a PMU-data-driven classification framework to distinguish these two disruptive events from the normal abrupt load changing in distribution networks. The rationale is that normal abrupt load changing has similar signatures on PMU data and advanced data analytics is required to distinguish the disruptive events from the normal load changing.

The classification input are six different features that have been extracted from PMU data as; change of voltage magnitude between two consecutive samples ($v(n+1)-v(n)$), change of voltage angle between two consecutive samples ($\delta_v(n+1)-\delta_v(n)$), current magnitude (p.u.), current angle (degree), change of current magnitude between two consecutive samples ($i(n+1)-i(n)$), and change of current angle between two consecutive samples ($\delta_i(n+1)-\delta_i(n)$). Moreover, since these features change over time, and we have a feature matrix, shown in (1), as opposed to a feature vector.

$$X = \begin{bmatrix} f_1^{(1)} & \cdots & f_1^{(p)} \\ \vdots & \ddots & \vdots \\ f_n^{(1)} & \cdots & f_n^{(p)} \end{bmatrix}_{n \times p} \quad (1)$$

where $X$ is the feature matrix and $f_i^{(t)}$ is the value of feature $i$ at time $t$.

In this paper, we consider the PMU data with two reporting rates as i) 60 samples per second (sps), e.g. SEL 651 [21], and ii) 120 sps, e.g. micro PMUs (μPMUs) developed at University of California, Berkeley [22]. Additionally, it is assumed that the capacitor bank switching takes about 1 cycle (16.67 *ms*) [23], and on-load tap changer switching takes about 30-200 *ms* [24] and the PMUs are capable of capturing this event.

In this paper, we present disruptive events classification using two different classification methods as; (1) PCA along with multi-class SVM, and (2) a neural network-based toolbox, i.e., autoencoder, along with a softmax layer classifier. Figure 1 illustrates the flowchart of these two methods which is discussed in detail in the next subsections.

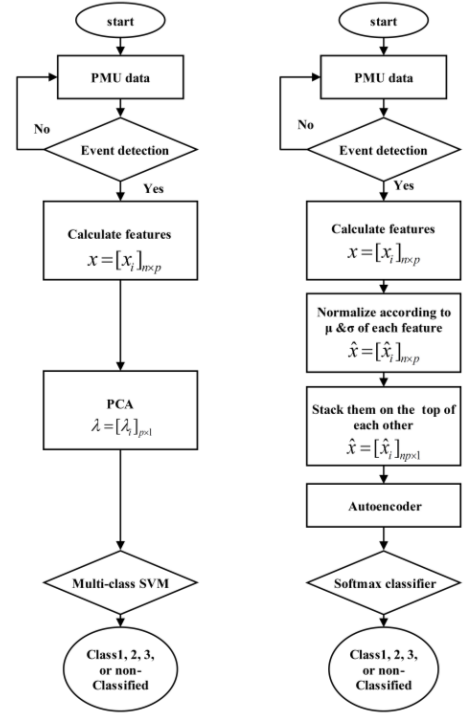

Fig. 1. a) PCA+SVM method (left), and b) Autoencoder+Softmax (right) events classification flowcharts

### A. PCA+SVM event classification algorithm

The extracted features from PMU data change over time, thus the features are presented as the feature matrix using (1). On the other hand the input to the SVM classifier is a vector and the matrices must be transformed into vectors. In this paper, PCA is utilized to obtain the dominant Eigen values of the feature matrix and used as the input to a multi-class SVM. The summary of PCA and SVM is presented below.

### A.1 Principal Component Analysis (PCA)

Principal component analysis (PCA) is a technique for reducing the dimensionality of the problem, and extracting the dominant features of the systems. In addition, it can be used in pattern recognition in high-dimensional data and it has a wide range of applications, e.g. image processing and face recognition, and data mining. For further study refer to [25].

### A.2 Support Vector Machine (SVM)

Support vector machine (SVM) is a supervised classification algorithm that uses linear or nonlinear hyper planes for separating classes from each other. The goal is to maximize the margin between the hyper planes, and therefore, the problem is formulated as a constrained optimization problem. Moreover, SVM can be applied to non-linear sets of data incorporating a method called kernel trick which maps original data into a higher-dimensional space. We used Gaussian kernel function in this paper. Additional discussion can be found in [26].

For multi-class classification, several algorithms have been proposed in the past. In this paper, a one-against-all algorithm is used to classify the events with respect to all the other classes using several binary SVMs.

*B. Autoencoder+ Softmax event classification algorithm*

As the second method, the event classification is carried out using a neural network-based toolbox, i.e., autoencoder, and softmax layer. In this method, the new feature matrix is first normalized using the mean and standard deviation of the historical feature matrices. The rows of the normalized feature matrix are then stacked on top of each other to create an input vector. The feature vector is then used as the input to the autoencoder layer for feature compression of the input vector. The softmax layer is then carried out using the compressed vector to classify the events. Fig. 1.b shows the flowchart and the summary is presented in the following.

*B.1 Autoencoder*

An autoencoder belongs to the artificial neural network family and is used as an unsupervised learning algorithm. It takes the data as input and tries to reconstruct the data through two layers of coding and decoding. The learning process is carried out using back propagation algorithms and the goal of training is to minimize the reconstruction error.

An autoencoder takes the $x \in R^d$ as input and maps it onto $z \in R^{d'}$ as
$$z = s_1(Wx + b) \quad (3)$$
Where $s_1$, $W$, and $b$ are the element-wise sigmoid function, the weight matrix, and the bias term, respectively. Then, $z$ is mapped onto the $R^d$ to reconstruct the input using
$$x' = s_2(W'z + b') \quad (4)$$
Where $s_2$, $W'$, and $b'$ are the element-wise sigmoid function, the weight matrix, and the bias term, respectively. [27].

*B.2 Softmax Classifier*

Softmax function is a generalization of logistic function that output a multiclass probability distribution as opposed to a binary probability distribution. It serves as the output layer of the autoencoder. It takes an m-vector x as input and outputs a y-vector of real values between 0 and 1. It is defined as
$$f_j(x) = \frac{e^{x_j}}{\sum_{i=1}^{m} e^{x_i}} \quad \text{for} \quad i=1,2,...m \quad (5)$$
Where $f_j(x)$ is the predicted probability for class $j$. Further discussion can be found in [28].

## III. SIMULATION AND RESULTS

In this paper, the proposed disruptive event classification is evaluated using the simulation studies. The PMU data for classification is generated by simulating the IEEE 13-node distribution system, as shown in Fig. 2. This distribution network has three different voltage levels as 115 kV, 4.16 kV, and 0.48 kV. The downstream network is connected via a 5000 kVA transformer to the upstream network. In this network there are (1) three single-phase regulators between bus 650 and bus 632, (2) a transformer between 633 bus and 634 bus, (3) a three-phase capacitor bank at bus 675, (4) a single-phase capacitor at bus 611, and (5) 15 distributed loads at different buses. We assume that one PMU is placed at bus 671 measuring voltage at this bus and current from bus 637 to bus 671. A Gaussian noise with zero mean and standard deviation of 1% of the measured values is then added to the voltage and current measurements (magnitudes and angles) to model the PMU inaccuracies.

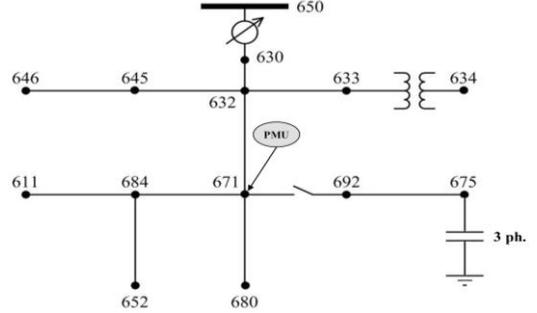

Fig. **2**. The IEEE 13 Node Test Feeder with one PMU [29]

Figure 3 shows the PMU voltage magnitude over one second, corresponding to three different classes, i.e. malfunctioned capacitor bank switching, malfunctioned OLTC switching of the voltage regulator and abrupt load changing. These figures demonstrate the voltage magnitude measurement of phase *a*.

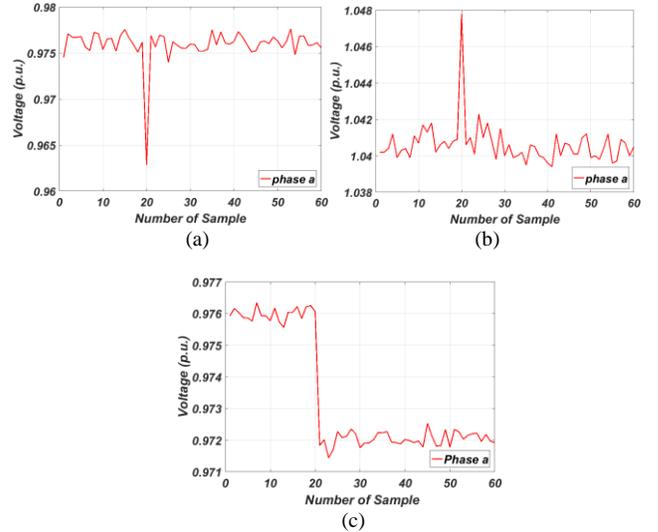

Fig. 3. PMU voltage magnitudes for three different classes a) malfunctioned capacitor bank switching, b) malfunctioned OLTC switching c) abrupt load changing

In order to create enough experiments for class 1, the malfunctioned three-phase capacitor bank switching event at bus 671 is simulated at different loading level. There are 15 different loads in the system, and for each of these loads, 10 different loading ranging from 50% up to 95%, with 5% intervals is considered. Therefore, 150 different experiments are simulated for class 1. Similarly, the same number of experiments is simulated for the malfunctioned OLTC switching event as the second class. Class three corresponds to normal abrupt load changing and it is assumed that one of the loads has a sudden change at a time. The abrupt load changing is simulated using 5%, 10%, 15%, 20%, and 25% increase or decrease of active and reactive power. Therefore, the total number of all class 3 experiments is 150, and we have generated 450 total number of experiments in all three classes.

The proposed multi-class classification algorithms are then trained and evaluated using the simulated PMU data. The classifiers are trained using $x$ percent of the data, i.e., selected randomly $x \in (10, 90)$, and the remaining data set is used to evaluate the classification accuracies as

$$Accuracy = \frac{number\ of\ accurate\ classification}{total\ number\ of\ test\ cases} \quad (6)$$

In this paper, we have gradually increased the percentage of the training data set and evaluated the confusion matrices and the classification accuracies. Table 1 and 2 demonstrate the confusion matrices for the scenario with 50% of data used for training and the rest for evaluation, using PCA+SVM and autoencoder+softmax, respectively.

Table 1. Confusion matrix in PCA+SVM method, with 50% of data used for training, and 60 sps PMU

|  |  | Predicted | | | |
| --- | --- | --- | --- | --- | --- |
|  |  | Class 1 | Class 2 | Class 3 | Non-classified |
| Actual | Class 1 | 53 (23.56%) | 7 (3.11%) | 12 (5.33%) | 3 (1.33%) |
|  | Class 2 | 8 (3.56%) | 54 (24%) | 6 (2.67%) | 7 (3.11%) |
|  | Class 3 | 10 (4.44%) | 2 (0.88%) | 62 (27.56%) | 1 (0.44%) |

Table 2. Confusion matrix in Autoencoder+Softmax method, with 50% of data used for training, and 60 sps PMU

|  |  | Predicted | | | |
| --- | --- | --- | --- | --- | --- |
|  |  | Class 1 | Class 2 | Class 3 | Non-classified |
| Actual | Class 1 | 63 (28%) | 6 (2.66%) | 6 (2.66%) | 0 (0%) |
|  | Class 2 | 9 (4%) | 59 (26.22%) | 7 (3.11%) | 0 (%) |
|  | Class 3 | 5 (2.22%) | 3 (1.33%) | 67 (29.77%) | 0 (%) |

As it is observed in the first row of Table 1, from 75 experiments corresponding to class 1, only 53 experiment are accurately classified, and 7, 12, and 3 experiments are misclassified as class 2, class 3, and non-classified, respectively. The percentage next to each number is the percentage of each number with respect to all the test cases.

The classification accuracy is then calculated using leave-one-out scenario which is a standard test for evaluation of any machine learning technique [30]. In this scenario it is assumed that all the experiments except one, i.e., 449 of the experiments, are used for training the classifiers and the only remaining experiment is tested using the trained classifiers. This process is carried out for the number of all experiments, i.e., 450 times, starting from the first experiment up to the last one. The accuracy is then calculated as the number of accurately classified experiment divided by the total number of experiments. The leave-one-out accuracies are 86.1% and 91.2% using PCA + SVM, and Autoencoder + Softmax, respectively which shows the better performance of the later method for disruptive event classification in distribution grids.

Finally the classification accuracies are calculated for different percentages of training cases, starting from 20% to 90%, with 10% intervals. Fig. 4 shows the results for PCA+SVM method for the two different sampling rates 60 sps and 120 sps. As it is observed from Fig. 4, the classification accuracies increase as more training data is used. Additionally, PMUs with 60 sps results in (a) 62% accuracy with 20% of data used for training, and (b) 84% accuracy with 90% of data used for training. While PMUs with 120 sps results in higher accuracies for the same percentage of training experiments, As the results verify, higher sampling rates leads to better capturing of the events, and consequently, better classification of the disruptive events.

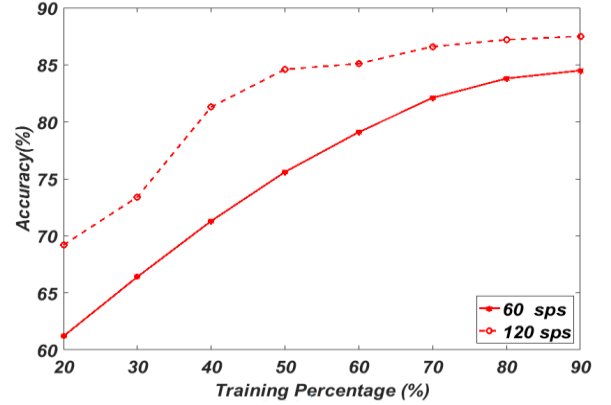
Fig. 4. Classification accuracy of PCA + SVM method for different training percentage and two different sampling rates

Fig. 5 shows the results using autoencoder+softmax method for two different sampling rates, i.e., 60 and 120 sps. Similar to PCA+SVM, as the training percentage increases, the accuracy increases. Additionally, as the sampling rate gets higher, the classification accuracies improve for the same percentage of the training cases. Furthermore, Figs. 4 and 5 are compared and it is verified that autoencoder+softmax method outperforms the PCA+SVM method in all different scenarios.

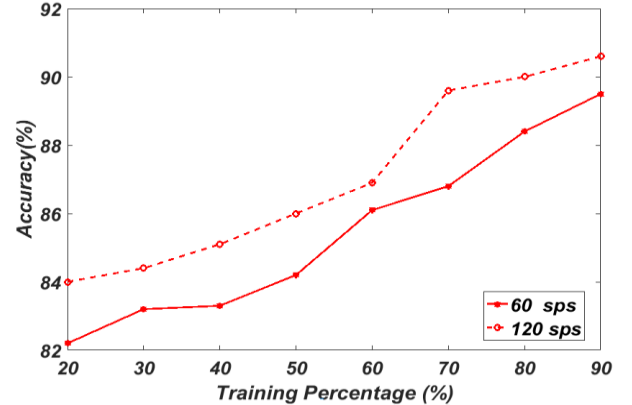
Fig. 5. Classification accuracy of autoencoder and softmax method for different training percentage and two different sampling rates

## IV. CONCLUSION AND FUTURE WORKS

Data-driven event detection in distribution grids provides essential operational and maintenance tools for next-generation smart grids with advanced measurement devices, e.g., micro-phasor measurement units (μPMUs). This paper presents a new framework for classification of disruptive events using PMU data in distribution grids. Two disruptive events are defined as malfunctioned capacitor bank switching and malfunctioned regulator on-load tap changer (OLTC) switching which provide similar PMU data pattern to normal abrupt load changes. The end result of this paper will provide a new framework that can be used for preventive maintenance scheduling of critical assets in distribution grids. In this paper, the event classification is developed using two multi-class classification algorithms for distinguishing between the two disruptive events and the normal load changing event. The first method is based on principal component analysis (PCA) along with multi-class support vector machine (SVM), and the second method is designed using autoencoder accompanied by the softmax layer classifier. The data for training and testing is provided using simulating the IEEE 13-bus distribution network with a PMU with 60 or 120 samples per second (SPS). The classification results verify the acceptable performance of the proposed framework for distinguishing the two disruptive events from the normal load change. The results also show the superiority of the second method over the first method. For future work, the authors will implement the proposed framework on larger standard networks, with more disruptive events, e.g. external voltage disturbance, voltage sag, reconfiguration. Furthermore, early-fusion, late-fusion techniques will be utilized for concatenation of several PMU data for wide-area disruptive event classification and localization frameworks.